\pdfoutput=1
\documentclass[11pt,a4paper]{article}
\usepackage{times,latexsym}
\usepackage{url}
\usepackage[T1]{fontenc}
\usepackage{graphicx}
\usepackage{subcaption}
\usepackage{booktabs, multirow} 
\usepackage{soul}
\usepackage{float}
\usepackage{microtype}
\usepackage[utf8]{inputenc}
\usepackage{microtype}
\usepackage[table]{xcolor}
\usepackage{verbatim}
\usepackage{listings}
\usepackage{float}
\usepackage{bbm}
\usepackage{amsmath}
\usepackage{arydshln}
\usepackage{pifont}
\newcommand{\cmark}{\ding{51}}%
\newcommand{\xmark}{\ding{55}}%

\definecolor{codegreen}{rgb}{0,0.6,0}
\definecolor{codegray}{rgb}{0.5,0.5,0.5}
\definecolor{codepurple}{rgb}{0.58,0,0.82}
\definecolor{backcolour}{rgb}{0.95,0.95,0.92}

\lstdefinestyle{codestyle}{
    backgroundcolor=\color{backcolour},   
    commentstyle=\color{codegreen},
    keywordstyle=\color{magenta},
    numberstyle=\tiny\color{codegray},
    stringstyle=\color{codepurple},
    basicstyle=\ttfamily\scriptsize,
    breakatwhitespace=false,         
    breaklines=true,                 
    captionpos=b,                    
    keepspaces=true,                 
    numbers=left,                    
    numbersep=5pt,                  
    showspaces=false,                
    showstringspaces=false,
    showtabs=false,                  
    tabsize=2
}

\lstdefinestyle{mystyle}{
  basicstyle=\ttfamily\footnotesize,
  columns=fullflexible,
  keepspaces=true,
  moredelim=[is][\bfseries\color{purple}]{[}{]}, 
  moredelim=[is][\textbf]{|}{|}, 
  moredelim=[is][\color{purple}]{@}{@}, 
  moredelim=[is][\bfseries\color{orange}]{&}{&}, 
}

\usepackage[acceptedWithA]{tacl2021v1}

\usepackage{xspace,mfirstuc,tabulary}



\newif\iftaclinstructions
\taclinstructionsfalse 
\iftaclinstructions
\renewcommand{\confidential}{}
\renewcommand{\anonsubtext}{(No author info supplied here, for consistency with
TACL-submission anonymization requirements)}
\newcommand{\instr}
\fi

\iftaclpubformat 

\else

\fi

\newif\ifcomments

\newcommand{\patrick}[1]{\ifcomments\textcolor{red}{\bf [{\sc PF:} #1]}\fi}

\newcommand{\gn}[1]{\ifcomments\textcolor{orange}{\bf [{\sc GN:} #1]}\fi}

\commentsfalse

\newcommand{\bison}{\textsc{Bison}}
\newcommand{\unicorn}{\textsc{Unicorn}}

\title{
The Devil is in the Errors: Leveraging Large Language Models for Fine-grained Machine Translation Evaluation
}

\author{
\textbf{Patrick Fernandes}\thanks{\:\,\, Work done while working part-time at Google.}$^{\:\,\,,2,3,4}$  \quad
\textbf{Daniel Deutsch}$^{1}$ \quad
\textbf{Mara Finkelstein}$^{1}$  \quad
\textbf{Parker Riley}$^{1}$  \quad \\
\textbf{André F. T. Martins}$^{3,4,5}$ \quad
\textbf{Graham Neubig}$^{2,6}$ \quad \\
\textbf{Ankush Garg}$^{1}$ \quad 
\textbf{Jonathan H. Clark}$^{1}$ \quad
\textbf{Markus Freitag}$^{1}$ \quad
\textbf{Orhan Firat}$^{1}$
\\
$^1$Google\quad 
$^2$Carnegie Mellon University \quad 
$^3$Instituto Superior Técnico  \quad \\
$^4$Instituto de Telecomunicações\quad 
$^5$Unbabel \quad 
$^6$Inspired Cognition \quad 
\\
{\small \texttt{pfernand@cs.cmu.edu}} 
}
\date{}

\begin{document}
\maketitle
\begin{abstract}
Automatic evaluation of machine translation (MT) is a critical tool driving the rapid iterative development of MT systems. While considerable progress has been made on estimating a single scalar quality score, current metrics lack the informativeness of more detailed schemes that annotate individual errors, such as Multidimensional Quality Metrics (MQM). In this paper, we help fill this gap by proposing \textbf{\textsc{AutoMQM}}, a prompting technique which leverages the \textit{reasoning} and \textit{in-context learning} capabilities of large language models (LLMs) and asks them to identify and categorize errors in translations. We start by evaluating recent LLMs, such as PaLM and PaLM-2, through simple \textit{score prediction} prompting, and we study the impact of labeled data through in-context learning and finetuning. We then evaluate \textsc{AutoMQM} with PaLM-2 models, and we find that it improves performance compared to just prompting for scores (with particularly large gains for larger models) while providing interpretability through error spans that align with human annotations. 
\end{abstract}

\section{Introduction}

\begin{figure}[t!]
  \centering
    \includegraphics[width=0.95\linewidth]{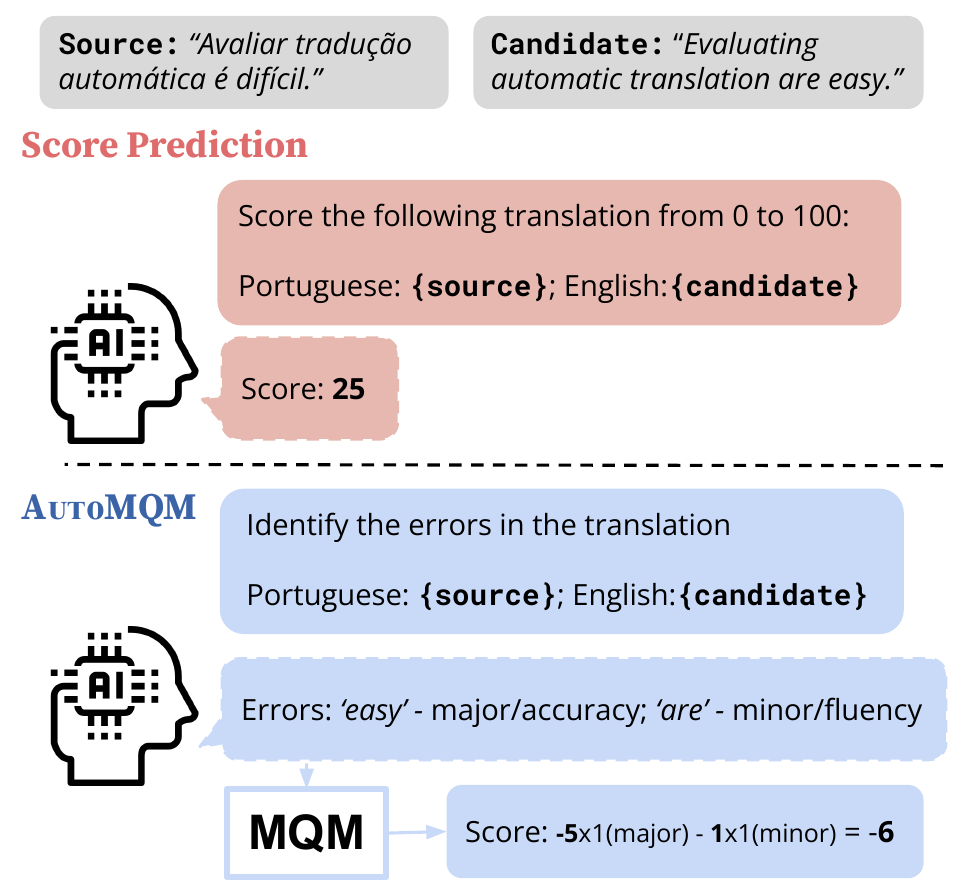}
  \caption{Illustration of how \textsc{AutoMQM} uses LLMs to assess the quality of a translation. Rather than asking for a single quality score, \textsc{AutoMQM} prompts models to identify and classify errors, and uses the MQM framework to produce a score.}
  \vspace{-0.5em}
  \label{fig:main-fig}
  \vspace{-1em}
\end{figure}

Evaluating natural language generation systems has always been challenging, and as the output quality of these systems has improved, evaluation has become even more challenging and critical.
For example, in Machine Translation (MT), a field where evaluation has garnered considerable attention, previous standard automatic surface-level metrics such as BLEU~\cite{papineni-etal-2002-bleu} are becoming less reliable as the quality of generation systems improves, with little remaining correlation with human judgments~\cite{freitag-etal-2022-results}.

To keep pace with the constantly improving quality of MT output, the next generation of automatic metrics is rapidly evolving.
\textit{Learned} automatic metrics that leverage human-judgments to finetune language models \cite{sellam-etal-2020-bleurt, rei-etal-2022-comet} currently represent the state-of-the-art in automatic evaluation benchmarks like the WMT Metrics task~\cite{freitag-etal-2022-results}, and show high correlation with human judgments. 
However, these metrics typically output a single, \textit{uninterpretable} quality score, making it difficult to understand the type and extent of errors identified by them. The lack of insights makes it difficult for model developers to leverage these metrics to improve their systems. \looseness=-1

Unlike automatic metrics that only provide a single scalar value as quality score, state-of-the-art human evaluation methodologies like Multidimensional Quality Metrics \citep[MQM;][]{lommel2014multidimensional, freitag-etal-2021-experts} ask professional annotators to identify and label error spans with a category and severity. This much richer feedback can be used to gain a better understanding of the current limitations of the model under evaluation and improve it. \looseness=-1

In this paper, we ask whether large language models (LLMs) in combination with a few human annotations can be used to design an automatic metric that generates rich feedback similar to that generated by human experts in MQM. 
This work is motivated by recent papers that demonstrated that LLMs can be used as automatic metrics \cite{liu2023geval} to generate a single quality score. In particular, \citet{kocmi2023large} showed that LLMs can be prompted to assess the quality of machine-generated translations, even achieving state-of-the-art performance on assessing system-level quality. However, previous work only provides a limited view of the capabilities of LLMs for machine translation evaluation: the focus has predominantly been on \textit{score prediction} (i.e. predicting a numerical value for quality), without considering the use of \textit{any} annotated data (either through in-context learning or finetuning), and only in \textit{high-resource} language pairs.

We provide a large-scale study of the capabilities of LLMs \citep[from the PaLM and PaLM-2 families; ][]{chowdhery2022palm,anil2023palm} for machine translation evaluation (both with and without a reference translation), provide a novel comparison between prompting and finetuning, and investigate the performance in the low-resource scenario. Inspired by findings that the performance of LLMs can be improved by prompting them for \textit{rationales} of their predictions \cite{wei2022chain, Lu2023EAPrompt}, we also propose \textbf{\textsc{AutoMQM}}, a prompting technique for MT evaluation that asks LLMs to identify error spans in a translation and to classify these errors according to the MQM framework, with a quality score derived automatically from the identified errors. A key advantage of \textsc{AutoMQM} is its \textit{interpretability}, as users can inspect the errors responsible for a score (\autoref{fig:main-fig}). 

Our contributions can be summarized as follows:
\begin{itemize}
    \itemsep0pt
    \item We confirm the finding of \newcite{kocmi2023large} that LLMs are \textit{zero-shot} state-of-the-art system-level evaluators, but show low correlation with human judgment compared to \textit{learned} metrics at the segment-level.
    \item We show that \textit{finetuning} an LLM with human judgment mitigates its low segment-level performance (particularly for smaller LLMs), showing similar correlations with human judgment at both the system-level and segment-level to state-of-the-art learned metrics.
    \item We are the first to evaluate LLM-based evaluation methods on low-resource language pairs. We find that their performance is promising, but lags behind state-of-the-art learned metrics. 
    \item We find that, with \textsc{AutoMQM}, PaLM-2 models can be prompted to generate rich MQM-like annotations, outperforming their score prediction counterparts at the segment-level. 
    \item Furthermore, annotations predicted by PaLM-2 models correctly identify over 50\% of words that are part of \textit{major} errors, and are comparable to the ones produced by state-of-the-art \textit{supervised} word-level evaluators.

\end{itemize}

Our findings might have significant implications for not only MT evaluation, but evaluation of machine-generated text in general, and further highlight the potential of using LLMs to provide \textit{AI Feedback} \cite{fernandes2023bridging}.

\section{Background: MT Evaluation}\label{sec:background}

Machine translation evaluation is one of the most well-studied evaluation problems in NLP \cite{ws-2008-statistical, freitag-etal-2022-results}. In this task, given 
\begin{enumerate}
    \itemsep0pt
    \item  a \textit{source} sentence in a (source) language
    \item  a \textit{candidate} translation in a (target) language
\end{enumerate}
an evaluation metric assesses the quality of the candidate translation by how well it conveys the meaning of the source sentence while considering other factors like \textit{fluency}. Like many other natural language generation evaluation problems, this task is difficult because the set of correct translations for a given source sentence is often very large and not entirely known in advance. To simplify the problem of machine translation evaluation, often (3) a \textit{reference} translation (typically created by a professional human translator) is included as additional information when assessing the candidate translation. This sub-problem is known as \textit{reference-based} evaluation (as opposed \textit{reference-less} evaluation or \textit{quality estimation}).

Up until recently, human evaluation of machine translation was carried out predominantly with the aim of assigning a single quality score to a candidate translation. Consequently, \textit{learned} metrics, which leverage collected human judgment data, are trained for and evaluated on the same task of \textit{score prediction} (i.e., assigning a single quality score to a candidate translation), and can achieve high correlation with human-provided scores \cite{freitag-etal-2022-results}.

However, framing machine translation evaluation as a score prediction task is problematic: any scoring or ranking of translations is implicitly based on an identification of errors in the candidate translations, and asking raters to solely provide a single score can lead to rushed and noisy judgments \cite{freitag-etal-2021-experts}. 

This insight has led to the adoption of the Multidimensional Quality Metrics (MQM) framework~\cite{lommel2014multidimensional, freitag-etal-2021-experts} as the gold standard for evaluating machine translation.
The MQM framework asks human evaluators to identify error spans in candidate translations and classify those errors according to various dimensions, e.g., \textit{fluency}, \textit{accuracy}, ... (see \autoref{app:mqm_description} for a more detailed description of MQM). Importantly, the MQM framework \textit{does not} ask annotators to provide a quality score for each translation, and instead derives one automatically from the identified error spans and their classifications. However, despite its richness, \textit{most} automatic metrics that leverage MQM data only use the final quality score produced by the framework and discard the error span information and classification.

\section{Related Work}
The success of \textit{learned} machine translation metrics \cite{sellam-etal-2020-bleurt, rei-etal-2022-comet, freitag-etal-2022-results, Qin2022T5ScoreDF}, which finetune neural network models pretrained on large amounts of (unsupervised) data, highlighted the importance of leveraging \textit{transfer learning} to achieve metrics with better correlation with human judgments. More recently, \textit{generative} LLMs \cite{openai2023gpt4, anil2023palm} have consistently demonstrated impressive results in natural language understanding and \textit{zero-} and \textit{few-shot} transfer and, naturally, interest in employing these models for (translation) evaluation has increased. \citet{kocmi2023large} first explored the use of GPT models for evaluating machine translation tasks, showing their potential as \textit{zero-shot} evaluators, and others have since extended GPT-based evaluation to other generation problems \cite{jain2023multidimensional, liu2023geval}. 

\citet{perrella-etal-2022-matese} first highlighted that MQM annotations could be leveraged to allow pretrained models to predict major and minor errors and, similarly to \textsc{AutoMQM}, used the identified errors to automatically score translations. However, their approach relied on weaker encoder-only or encoder-decoder language models, required \textit{supervised} data to work, and overall underperformed other top metrics. We compare against their \textit{MaTASe} metric in our experiments.
\citet{Lu2023EAPrompt} showed that doing \textit{error analysis}, a prompting technique similar to \textsc{AutoMQM}, could lead to better ChatGPT-based evaluators. However, they still relied on the LLM to provide a score once it identified errors (rather than do it automatically using something like the MQM framework). Furthermore, they provided a very limited meta-evaluation using only 40 examples per language pair. Concurrently with our work, \citet{xu2023instructscore} proposed \textsc{InstructScore}, a LLaMA-based evaluator that asks models to identify and categorize errors in translation (as well as providing a natural language explanation for each error). However, the authors only explore a 7B parameter model and don't leverage zero- and few-shot capabilities of models as in this work. Instead, they rely on a more complex approach of distilling the knowledge of a more capable GPT-4 LLM. 

Additionally, WMT Word-Level Quality Estimation shared tasks \cite{fonseca-EtAl:2019:WMT, zerva-etal-2022-findings} leverage MQM data by converting span-level annotations of errors (normally of \textit{major} severity) to word-level tags and Task 2 in the WMT19 Quality Estimation shared task evaluation explicitly evaluated submissions of span-level annotations (although most submissions still consisted of models that predicted word-level tags which were converted to spans). We also compare against state-of-the-art word-level quality estimation models.

\section{Using LLMs to Predict Quality Scores}\label{sec:score_prediction}

Recent works have shown that large language models are versatile, general-purpose models that can be used to tackle many problems in NLP, including evaluation \cite{kocmi2023large, jain2023multidimensional, liu2023geval}. We begin by exploring how LLMs can be used for machine translation evaluation through \textit{score prediction}.

\subsection{Prompting}

We start by measuring how far we can push the performance of LLMs with just \textit{prompting} \cite{liu2023prompt}: by defining the task of MT evaluation and quality estimation as \textit{textual templates} (with a general description of the problem and ``slots'' for the inputs and outputs), we can use general-purpose LLMs to perform these tasks at inference-time, without any parameter updates. 

Throughout the paper, we choose to use \citet{kocmi2023large}'s \texttt{GEMBA-SQM} prompt (\autoref{fig:score_prediction_prompt}), which asks models to generate (a string representation of) a score from 0-100. We choose this prompt for two reasons: firstly, early explorations with theirs and other prompts\patrick{stretch goal: @Mara's explorations with other prompts}\gn{Might want a citation for ``other prompts''} showed that this generally performed well. Secondly, using a single prompt ensures a fairer comparison between the capabilities of different models.\footnote{While this prompt wasn't the best for \textit{system-level}, it led to the best \textit{segment-level} performance in GEMBA.}

\begin{figure}[ht]
\centering
\begin{lstlisting}[style=mystyle]
Score the following translation from 
|{src_lang}| to |{tgt_lang}| @with respect 
to the human reference@ on a continuous 
scale from 0 to 100 that starts with 
"No meaning preserved", goes through 
"Some meaning preserved", then "Most 
meaning preserved and few grammar mistakes",
up to "Perfect meaning and grammar".

|{src_lang}| source: "|{source}|"
[{tgt_lang}] @human reference:@ ["{reference}"]
|{tgt_lang}| translation: "|{candidate}|"
Score (0-100): &{score}&
\end{lstlisting}
\vspace{-0.8em}
\caption{\small The \textit{score prediction} prompt used in this paper. Equivalent to the \texttt{GEMBA-SQM} prompt in \citet{kocmi2023large}. Parts in  {\color{purple} purple} are only included for \textit{reference-based} evaluation, while parts in {\color{orange} orange} represent slots for outputs and are only included for in-context examples.}
\label{fig:score_prediction_prompt}
\vspace{-1.5em}
\end{figure}

\paragraph{In-Context Learning} A surprising emergent capability of LLMs is their ability to improve on prompting-based tasks by including a very small amount of labeled data as part of the prompt/context \cite{NEURIPS2020_1457c0d6} and \textit{without} parameter updates, a technique called \textit{in-context learning} (ICL). We thus investigate the impact that ICL has on LLMs' ability to assess translation quality. Recent works have shown that the impact of ICL is tightly tied with the exact examples included in the prompt, with a poor selection procedure leading to no improvements or even worse performance than the zero-shot case \cite{jain2023multidimensional}. We therefore explore two sampling approaches to select in-context examples from a pre-defined ``pool'' of translation quality assessments: \textbf{uniform sampling} and \textbf{stratified sampling}, where the example pool is bucketed by score ranges and examples are sampled from each bucket.

\subsection{Finetuning}
\label{subsec:finetuning}
It has previously been shown that LLMs are capable of zero-shot evaluation \cite{kocmi2023large}, but the extent to which \textit{finetuning} on human judgment data can further boost the performance of LLMs has not been studied. In the WMT'22 Metrics Shared Task \citep{freitag-etal-2022-results}, all top submissions were learned metrics; that is, pretrained models finetuned on human judgment data\footnote{While these metrics all leverage powerful pretrained (language) models, these generally aren't considered LLMs}.

\begin{figure*}[ht]
    \centering
\begin{lstlisting}[style=mystyle]
Based on the given source and reference, identify the major and minor errors in this
translation. Note that Major errors refer to actual translation or grammatical errors,
and Minor errors refer to smaller imperfections, and purely subjective opinions about 
the translation.

|{src_lang}| source: "|{source}|"
[{tgt_lang}] @human reference:@ ["{reference}"]
|{tgt_lang}| translation: "|{candidate}|"
Errors: &{error1:span} - {error1:severity}/{error1:category}; {error2:span} - ... &
\end{lstlisting}
\vspace{-0.7em}
\caption{The \textit{\textsc{AutoMQM}} prompt used in this paper. {\color{purple} Parts in purple} are only included for \textit{reference-based} evaluation, while  {\color{orange} parts in orange} represent slots for outputs, and are only included for in-context examples.}
\label{fig:automqm_prompt}
\vspace{-0.5em}
\end{figure*}

Thus, we investigate whether LLMs are amenable to finetuning on human judgment data. LLMs used in top-performing metrics are generally much larger than the pretrained language models leveraged by previous learned metrics (which generally have fewer than 1 billion parameters). Moreover, most learned metrics leverage pretrained encoder-only rather than (decoder-only) prefix language models.
\gn{Why is this an interesting/important distinction? Is it because it makes it less clear whether finetuning would work? It would be nice to mention this briefly. Also, T5Score used T5-XXLarge with 11B parameters.}
We experiment with finetuning LLMs using two objectives:
\begin{itemize}
    \itemsep0em
\item \textit{Regression} (\textbf{R}): Commonly used for training learned metrics \cite{rei-etal-2022-comet}, the objective here is a regression loss (e.g., mean squared error) between continuous scores obtained from the model (for example, with a \textit{regression head}) and the human scores.
\item \textit{Generative Classification} (\textbf{GC}): We bucket scores into discrete classes (see \S\ref{subsec:experimental-setup}) and treat the MT evaluation task as a text-to-text classification problem \cite{colin20t5}.
\end{itemize}

\section{Using LLMs to Predict Error Spans}

While producing quality scores that correlate with human judgments is an important part of translation quality assessment, metrics that solely do score prediction suffer from problems of \textbf{interpretability}: if a metric assigns a low score, the downstream users are left in the dark about which parts of the translation were responsible for the score and thus need to be corrected. This is especially problematic in cases where the metric assigns a \textit{wrong} score to a translation, as it is much harder to diagnose why the evaluation model made a mistake, and identify and prevent similar mistakes in the future. In fact, reducing translation quality to a single score has proven problematic even for human annotators: asking raters to solely provide a single score can lead to rushed and noisy judgments \cite{freitag-etal-2021-experts} and the current gold standard for translation quality evaluation involving human annotators is instead based on methodologies like the MQM framework (see \S\ref{sec:background}) , which provide richer feedback by identifying error spans, categorizing them, and evaluating their severity.

Interestingly, another emergent phenomenon in LLMs is the success of \textit{chain-of-thought} prompting \cite{wei2022chain}: when defining a prompt for a particular task, if we instruct the model to produce a series of intermediate reasoning steps (\textit{``let's think step-by-step''}), it tends to generate a free-text \textit{rationale} before generating an output, and this often improves the performance on the task at hand \cite{liu2023geval}. Furthermore, this \textit{chain-of-thought} prompting can be used to obtain \textit{structured} rationales from LLMs, and this can lead to better performance than with free-text rationales \cite{Lu2023EAPrompt}.

Motivated by these findings, we propose \textbf{\textsc{AutoMQM}}, a prompting technique for translation quality assessment that instructs LLMs to \textit{identify} errors in a translation, and \textit{categorize} the type of error according to the MQM framework \cite{lommel2014multidimensional}. Furthermore, we \textit{don't} ask the model to produce a score, as the MQM framework provides an algorithmic procedure to obtain one from identified errors: the total score is the sum of penalties for all errors identified, where (roughly) \textit{major} errors get penalized with $-5$ and \textit{minors} with $-1$ (see \autoref{app:mqm_description} for a more detailed description of the scoring algorithm).\footnote{This is similar to methods that leverage external \textit{executors} to improve the performance of LLMs \cite{gao2022pal}} \autoref{fig:automqm_prompt} shows the main \textsc{AutoMQM} prompt used in this paper.

Importantly, obtaining meaningful \textsc{AutoMQM} results in a zero-shot setting is a substantially more challenging task compared to score prediction: we found that, without any in-context examples, LLMs tend to produce outputs that are either uninformative or difficult to parse. Thus we only consider the \textsc{AutoMQM} task in the \textit{few-shot} scenario. Based on the findings from \S\ref{subsec:score_prediction_results}, we explore the impact of in-context learning by sampling from the example pool using stratified sampling extended with a set of \textit{rejection criteria} (\autoref{app:rejection-criteria}), which ensures that the example set has a balance between major and minor errors as well as diversity in the categories of errors.

\section{Experiments}

\subsection{Experimental Setup}
\label{subsec:experimental-setup}

\begin{table}[t]
    \centering
\footnotesize
    \begin{tabular}{ccr}
\toprule
        \bf LP & \bf \#Sys & \bf \#Seg \\
        \midrule
        en$\rightarrow$de & 13 & 1315 \\
        zh$\rightarrow$en & 14 & 1875 \\
        en$\rightarrow$ru & 15 & 1315 \\
\bottomrule
\end{tabular}
\quad
\begin{tabular}{ccr}
\toprule
        \bf LP & \bf \#Sys & \bf \#Seg \\
        \midrule
        en$\rightarrow$kk & 11 & 998 \\
        kk$\rightarrow$en & 11 & 1000 \\
        en$\rightarrow$gu & 11 & 998 \\
        gu$\rightarrow$en & 11 & 1016 \\
        \bottomrule
\end{tabular}
    \vspace{-0.7em}
    \caption{\small The number of systems and segments that have MQM scores (left) and DA scores (right)
    used as ground-truth in this work.}
    \label{tab:dataset_stats}
    \vspace{-1em}
\end{table}

\paragraph{Data}
The metrics in this work are evaluated on both \textit{high-resource} and \textit{low-resource} language pairs.
The three high-resource language pairs come from the WMT'22 Metrics Shared Task \citep{freitag-etal-2022-results}: en$\rightarrow$de, zh$\rightarrow$en, and en$\rightarrow$ru.
The ground-truth translation quality scores are derived from MQM ratings in which expert annotators marked error spans in the translations with different severity levels which are automatically converted to a numeric score (see \S\ref{sec:background}).
The four low-resource language pairs come from the WMT'19 Metrics Shared Task \citep{ma-etal-2019-results}: en$\leftrightarrow$gu and en$\leftrightarrow$kk.
Since MQM ratings are not available for the low-resource pairs, the ground truth quality scores are direct assessment (DA) scores.
DA scores are quality assessments assigned by non-expert raters on a scale from 0-100, then normalized per rater.
See Table~\ref{tab:dataset_stats} for statistics about the number of MT systems and segments for every language pair. 

Additionally, in our experiments, \textsc{AutoMQM} required in-context examples with MQM annotations to work, so we restrict our evaluation of \textsc{AutoMQM} to en$\rightarrow$de and zh$\rightarrow$en because there are available MQM ratings from the WMT'21 Metrics Shared Task \cite{freitag-etal-2021-results} that we can use as in-context learning example pools.

\paragraph{Models}
We base most of our experiments on the following LLMs:

\begin{itemize}
    \item \textbf{PaLM}: A 540 billion parameter autoregressive Transformer model trained on 780 billion tokens of high-quality text \cite{chowdhery2022palm}. It showed remarkable performance on a wide-range of NLP tasks, including Machine Translation \cite{vilar2022prompting}.
    \item \textbf{PaLM-2}: The successor to PaLM, the PaLM-2 family of LLMs \cite{anil2023palm} builds upon recent research insights, such as compute-optimal scaling, a more multilingual and diverse pre-training mixture, and architectural/optimization improvements. We mainly use two model sizes in the family: PaLM-2 \bison{} and (the larger) PaLM-2-\unicorn{}.\footnote{Information about exact number of parameters of PaLM-2 models is not publicly available.}
    In addition we explore the impact of instruction-tuning by using a \unicorn{} model finetuned on the FLAN dataset \cite{weifinetuned}. 
\end{itemize}

For \textit{score prediction}, we compare PaLM and PaLM-2 against the GPT family of LLMs \cite{NEURIPS2020_1457c0d6, openai2023gpt4} by leveraging the results and outputs from the GEMBA evaluator \cite{kocmi2023large}.  We then evaluate the performance of \textsc{AutoMQM} with only PaLM-2 models (which performed best in score prediction).

Additionally, for the high-resource languages, we compare to a set of strong baseline evaluation metrics, MetricX-XXL and COMET-22, which were the two top-performing metrics in the WMT'22 Metrics Shared Task.
MetricX-XXL and COMET-22 are both finetuned regression models trained on DA data from WMT that are initialized with mT5 \citep{xue-etal-2021-mt5} and XLM-R \citep{conneau-etal-2020-unsupervised}, respectively. 

For the \textsc{AutoMQM} experiments, we also compare against \textsc{MaTESe}, a comparable submission to the WMT'22 Metrics Shared task that finetuned a XLM-R model to identify major and minor errors, and computed a score automatically. Since we were unable to obtain the span-level predictions for the \textsc{MaTESe} submission, we also compare against the top submission to the WMT'22 Word-Level Quality Estimation Shared Task \cite{zerva-etal-2021-ist}: word-level \textsc{CometKiwi} (COMET-WL) \cite{rei-etal-2022-cometkiwi}, also based on an XLM-R model trained on a combination of sentence- and word-level data. To do so, we re-run this model on the WMT'22 Metrics Shared Task data, and convert the predicted \textit{word-level} \texttt{OK/BAD} tags into spans.\footnote{We consider a span as any maximal consecutive sequence of words marked as \texttt{BAD}, assigning every span the \textit{major} severity.}


\paragraph{Finetuning} For \textit{regression} finetuning, we use a real-valued logit, extracted from a fixed index in the first target token's logit vector, as the quality signal. (In particular, we leverage a special, \textit{unused}, vocabulary token.) This was the technique used to train MetricX-XXL in the WMT 2022 Shared Task submission \citep{freitag-etal-2022-results}. The regression-based model was trained on WMT direct assessment (DA) data from the years 2015 through 2020.

For \textit{generative} classification, we bucket the scores in the training data into five classes, where class boundaries are assigned so that each class contains an equal number of training examples. We then map labels to verbal ratings from the following set, based on their bucket: [\textit{"very bad"}, \textit{"bad"}, \textit{"ok"}, \textit{"good"}, \textit{"very good"}]. To evaluate the model, predictions are mapped back to integer labels from 1 to 5. Any predictions not containing a substring in the label set are considered invalid and are mapped to 0. We experimented with finetuning on both DA and MQM 2020 \citep{freitag-etal-2021-experts} data, and found that the latter performed slightly better.

To assess the impact of \textit{model size}, we also finetune two additional (smaller) PaLM-2 models, which we call $S$ and $M$, comparing their finetuned and zero-shot performance.\footnote{We use a small variation of the \textit{zero-shot} prompt, asking models for scores from the same 5 buckets used in finetuning.}

\paragraph{Metric Meta-Evaluation}

The quality of an automatic evaluation metric is estimated by comparing the agreement between the metric scores and ground-truth quality scores on a large number of translations from different MT systems, a process known as metric meta-evaluation.
This work reports three different agreement scores, as follows.

The first is system-level accuracy, which calculates the percent of system pairs that are ranked the same by the metric and ground-truth scores, micro-averaged over a set of language pairs \citep{kocmi-etal-2021-ship}.
System-level scores are defined as the average score across all segments.

At the segment-level, the standard correlation that is reported by WMT is Kendall's $\tau$.
However, recent work pointed out problems with Kendall's $\tau$ with respect to ties \citep{deutsch2023ties}.
In short, different variants of $\tau$ are inconsistent with respect to ties and even biased against metrics that predict ties, as our metrics do in this work.
\citet{deutsch2023ties} recommend reporting a pairwise accuracy score, which rewards metrics for correctly ranking translations as well as correctly predicting ties, in combination with a tie calibration procedure that automatically introduces ties into metric scores so that the meta-evaluation is fairer.
This accuracy score, denoted acc$^*$, ranges between 0 and 1, and a random metric would achieve 33\% accuracy.
We report the ``group-by-item'' variant of the pairwise accuracy score from \citet{deutsch2023ties} in addition to Pearson's $\rho$, a complementary signal to rank-based correlations that measure the strength of the linear relationship between two variables (and one of the standard correlations reported in WMT). \looseness=-1

\begin{table*}[ht]
\centering
\footnotesize
{\renewcommand{\arraystretch}{0.75}
\begin{tabular}{@{\extracolsep{1pt}}lcccccccc}
\toprule
 &  & \multicolumn{1}{c}{\textbf{System-Level}} & \multicolumn{6}{c}{\textbf{Segment-Level}} \\
\cmidrule{3-3}\cmidrule{4-9}
 & & All (3 LPs) & \multicolumn{2}{c}{EN-DE} & \multicolumn{2}{c}{ZH-EN} & \multicolumn{2}{c}{EN-RU} \\
\cmidrule{3-3}\cmidrule{4-5} \cmidrule{6-7} \cmidrule{8-9}
Model & Ref? & Accuracy & $\rho$ & $\mathrm{acc}^\star$ & $\rho$ & $\mathrm{acc}^\star$ & $\rho$ & $\mathrm{acc}^\star$ \\
\midrule
\multicolumn{1}{@{}l}{\raisebox{1.2ex}{\scriptsize \textbf{Baselines}}}\\[-6pt]
MetricX-XXL & \cmark &  85.0\% & 0.549 & 61.1\% & 0.581 & 54.6\% & 0.495 & 60.6\% \\
COMET-22 & \cmark &  83.9\% & 0.512 & 60.2\% & 0.585 & 54.1\% & 0.469 & 57.7\% \\
\rowcolor{gray!8} COMET-QE & \xmark &  78.1\% & 0.419 & 56.3\% & 0.505 & 48.8\% & 0.439 & 53.4\% \\
\midrule
\multicolumn{1}{@{}l}{\raisebox{1.2ex}{\scriptsize \textbf{Prompting}}}\\[-6pt]
PaLM 540B & \cmark &  90.1\% & 0.247 & 55.4\% & 0.255 & 48.5\% & 0.180 & 48.6\% \\
PaLM-2 \bison{} & \cmark &  88.7\% & 0.394 & 56.8\% & 0.322 & 49.3\% & 0.322 & 52.8\% \\
PaLM-2 \unicorn{} & \cmark &  90.1\% & 0.401 & 56.3\% & 0.349 & 51.1\% & 0.352 & 55.3\% \\
FLAN-PaLM-2 \unicorn{} & \cmark & 75.9\% & 0.197 & 55.6\% & 0.139 & 46.1\% & 0.198 & 52.0\% \\
\rowcolor{gray!8} PaLM 540B & \xmark &  84.3\% & 0.239 & 56.1\% & 0.270 & 43.1\% & 0.300 & 51.8\% \\
\rowcolor{gray!8} PaLM-2 \bison{} & \xmark & 85.0\% & 0.355 & 57.0\% & 0.299 & 48.6\% & 0.303 & 53.1\% \\
\rowcolor{gray!8} PaLM-2 \unicorn{} & \xmark &  84.3\% & 0.275 & 56.1\% & 0.252 & 48.3\% & 0.209 & 49.8\% \\
\rowcolor{gray!8} FLAN-PaLM-2 \unicorn{} & \xmark & 69.7\% & 0.116 & 54.6\% & 0.112 & 43.8\% & 0.156 & 47.8\% \\
\midrule
\multicolumn{1}{@{}l}{\raisebox{1.2ex}{\scriptsize \textbf{Finetune}}}\\[-6pt]
PaLM-2 \bison{} (\textbf{R})  & \cmark &  88.0\% & 0.511 & 61.0\% & 0.459 & 51.5\% & 0.458 & 59.5\% \\
PaLM-2 \bison{} (\textbf{GC})  & \cmark &  86.1\% & 0.400 & 59.2\% & 0.444 & 49.3\% & 0.365 & 56.0\% \\
PaLM-2 \unicorn{} (\textbf{R})  & \cmark &  87.6\% & 0.508 & 61.1\% & 0.412 & 52.6\% & 0.460 & 60.4\% \\
\rowcolor{gray!8} PaLM 2 \bison{} (\textbf{R}) & \xmark &  87.6\% & 0.490 & 59.9\% & 0.439 & 53.4\% & 0.437 & 59.2\% \\
\rowcolor{gray!8} PaLM 2 \bison{} (\textbf{GC}) & \xmark &  86.1\% & 0.368 & 57.5\% & 0.420 & 47.3\% & 0.390 & 54.9\% \\
\rowcolor{gray!8} PaLM 2 \unicorn{} (\textbf{GC}) & \xmark &  86.1\% & 0.407 & 57.9\% & 0.402 & 45.6\% & 0.411 & 55.3\% \\
\bottomrule
\end{tabular}
\vspace{-0.5em}
\caption{Meta-evaluation results at system and segment-level for the \textit{high-resource} language pairs. 
Finetuned \textbf{(R)} and \textbf{(GC)} represent the \textit{regression} and \textit{generative classification} objectives (\S\ref{subsec:finetuning}).
\cmark and \xmark\; represent \textit{reference-based} and \textit{reference-less} metrics, respectively. 
}
\label{tab:score-prediction}
}
\vspace{-1em}
\end{table*} 


\paragraph{Span Meta-Evaluation} Since \textsc{AutoMQM} provides not only scores but also the identified error spans, we can compare the predicted spans with the errors marked by annotators in the MQM annotations. We evaluate quality of predicted spans using: (1) \textit{Span Precision} (SP), which measures the overlap of predicted spans and gold (annotated) spans; and (2) \textit{Major recall} (MR), which captures the percentage of gold major errors that were predicted as errors (either minor or major).

More formally, consider the set of ground truth spans $S^\star$, where each span consists of a sequence of words, i.e., $s_i = (w_{(a)}, w_{(a+1)}, \cdots)$. Let $S^\star_{\mathrm{maj}} \subseteq S^\star$ be the subset containing only the major errors. Given a span set $S$, we define its positional set $P(S)$ as the set containing the positions of all the words in every span in $S$. For example, assuming a span $s_i = (w_{(n)}, w_{(n+1)}, \cdots)$ in $S$ starts at the $n$th position in the text, its corresponding positional set will include the positions $\{n, n+1, ..., n+\mathrm{len}(s_i)-1\}$. Then for a set of \textit{predicted} spans $\hat{S}$, SP and MR are defined as:
\begin{align}
\mathrm{SP}(\hat{S}) &= \frac{|P(\hat{S}) \cap P(S^\star)|}{|P(\hat{S})|}  \\
\mathrm{MR}(\hat{S}) &= \frac{|P(\hat{S}) \cap P(S^\star_{\mathrm{maj}})|}{|P(S^\star_{\mathrm{maj}})|}
\end{align}


Intuitively, we care for overall precision (regardless of severity) since we want to make sure predicted errors tend to be marked by annotators as well, but for recall we care mostly for \textit{major} errors, as these have a larger impact on translation quality and are more critical to identify.  Additionally, we also report the (3) \textit{Matthews Correlation Coefficient} (MCC), one of the official metrics in the word-level quality estimation tasks \cite{zerva-etal-2022-findings}. \looseness=-1

\subsection{Results}
\label{subsec:score_prediction_results}

\subsubsection{Score Prediction}

\autoref{tab:score-prediction} summarizes the meta-evaluation results, at the \textit{system} and \textit{segment} level, for both the \textit{zero-shot prompting} and \textit{finetuning} settings. 

\paragraph{Prompting} A first observation is almost all zero-shot LLM evaluators have higher \textit{system-level} performance than learned metrics (with and without references), with PaLM 540B and PaLM-2 \unicorn{} achieving the best performance. At the segment level, the story is more complicated: similarly to \citet{kocmi-etal-2022-findings}, we find that none of the LLMs we explored was able to consistently outperform the baseline learned metrics. We see that PaLM-540B is a particularly poor reference-based evaluator, which is surprising given its system-level performance. Unexpectedly, instruction-tuning with FLAN seems to \textit{degrade} performance, with FLAN-PaLM-2 \unicorn{} achieving poor performance at both the system and segment levels.\footnote{Note that this might be a problem with the FLAN dataset and not instruction-tuning in general, as the GPT models are also instruction-tuned and perform well.}

\begin{table}[ht]
\centering
\small
\setlength{\tabcolsep}{1.5pt}
\renewcommand{\arraystretch}{0.83}
\begin{tabular}{@{\extracolsep{0.1pt}}lcccccc}
\toprule
 &  & \multicolumn{1}{c}{\textbf{System}} & \multicolumn{3}{c}{\textbf{Segment} $\mathrm{acc}^\star$} \\
\cmidrule{3-3}\cmidrule{4-6}
Model & Ref? & All & {\small EN-DE} & {\small ZH-EN} & {\small EN-RU} \\
\midrule
\multicolumn{1}{@{}l}{\raisebox{1.2ex}{\scriptsize \textbf{GEMBA}}}\\[-6pt]
GPT-3.5 & \cmark & 85.4\% & 54.9\% & 49.5\% & 47.5\% \\
GPT-4 & \cmark & 88.7\% & 57.8\% & 52.6\% & 55.0\% \\
\rowcolor{gray!8} GPT-3.5 & \xmark & 82.5\% & 56.1\% & 49.7\% & 49.3\% \\
\rowcolor{gray!8} GPT-4 & \xmark & 89.1\% & 56.4\% & 53.4\% & 54.8\% \\
\midrule
\bison{} & \cmark & 88.7\% & 56.8\% & 49.3\% & 52.8\% \\
\unicorn{} & \cmark & 90.1\% & 56.3\% & 51.1\% & 55.3\% \\
\rowcolor{gray!8}\bison{} & \xmark & 85.0\% & 57.0\% & 48.6\% & 53.1\% \\
\rowcolor{gray!8}\unicorn{} & \xmark & 84.3\% & 56.1\% & 48.3\% & 49.8\% \\
\bottomrule
\end{tabular}
\vspace{-0.7em}
\caption{\small Comparison between PaLM-2 and GPT-based GEMBA \cite{kocmi-etal-2022-findings} at the system and segment levels for the \textit{high-resource} language pairs.}
\label{tab:gemba}
\vspace{-1em}
\end{table} 

Nevertheless, PaLM-2 models achieve high correlations with human judgments, and the \textit{reference-less} PaLM-2 \bison{} is competitive with the \textit{learned} baselines, particularly at assessing alternative translations of the same sentence ($\mathrm{acc}^*$). When comparing PaLM-2 models with \citet{kocmi-etal-2022-findings}'s GPT-based GEMBA evaluator (\autoref{tab:gemba}), we see that both families of LLMs perform similarly, with PaLM-2 models exhibiting higher system-level performance than GPT-based GEMBA, while GEMBA achieves better segment-level accuracy, particularly in the reference-less setting.

\begin{figure}[ht]
  \centering
    \includegraphics[width=0.92\linewidth]{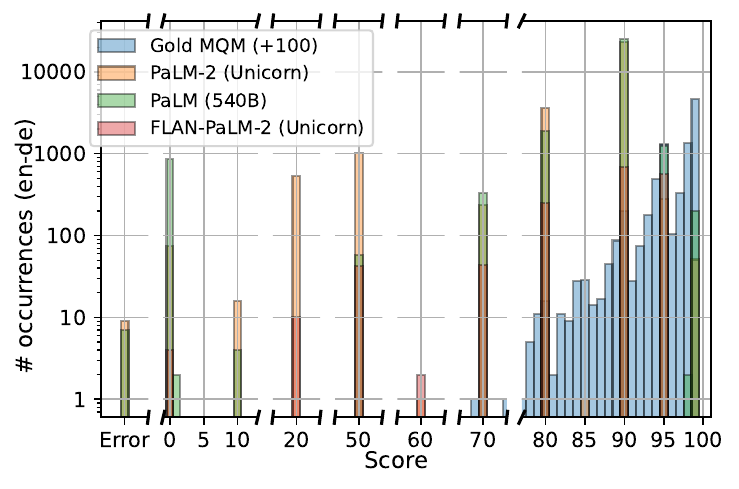}
  \vspace{-1em}
  \caption{\small Distribution of scores for various LLM \textit{reference-based} evaluators, on the EN-DE test set. Note that the $y$ axis is in \textit{log-scale}.}
  \label{fig:distribution_of_scores}
  \vspace{-1.2em}
\end{figure}

\autoref{fig:distribution_of_scores} shows the distribution of scores produced by PaLM- and PaLM-2-based evaluators. We find that, despite being prompted to give a score in the 0-100 range, these models almost always output one of a very limited set of scores (e.g. 0, 50, 90, 95). Given \citet{kocmi2023large}'s similar findings with GPT models, it seems that this is a consequence of the pretraining objective.

\begin{figure}[ht]
  \centering
  \hspace{-2em}
    \includegraphics[width=0.85\linewidth]{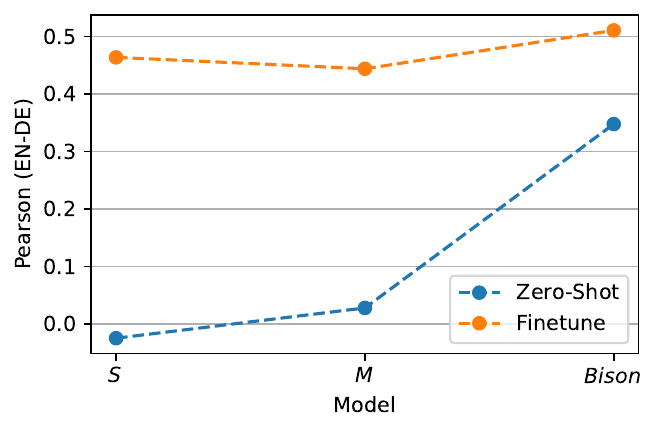}
  \vspace{-1em}
  \caption{\small Behavior of \textit{Pearson} as we scale the LLM's parameter count. Note that the $x$ axis is not to-scale with regard to parameter count.}
  \label{fig:scaling_model}
  \vspace{-0.5em}
\end{figure}

\paragraph{Finetuning}
Despite their already-great performance in the zero-shot setting, we find that finetuning LLMs can further improve LLM evaluators' segment-level scores. This is particularly obvious for the \textit{reference-less} evaluators, where a finetuned PaLM-2 \bison{} achieves state-of-the-art performance in segment-level correlations and comparable system-level accuracy across all language pairs. \patrick{comment on classification vs regression}
Moreover, when we look at how performance \textit{scales} with parameter count (\autoref{fig:scaling_model}), we observe an interesting trend: while smaller models are not capable of being effective zero-shot evaluators, finetuning them leads to competitive performance, and only a slight decrease when compared to their larger finetuned counterparts.

\paragraph{In-context Learning}

\begin{figure}[ht]
    \centering
      \hspace{-1.5em}
    \includegraphics[width=0.85\linewidth]{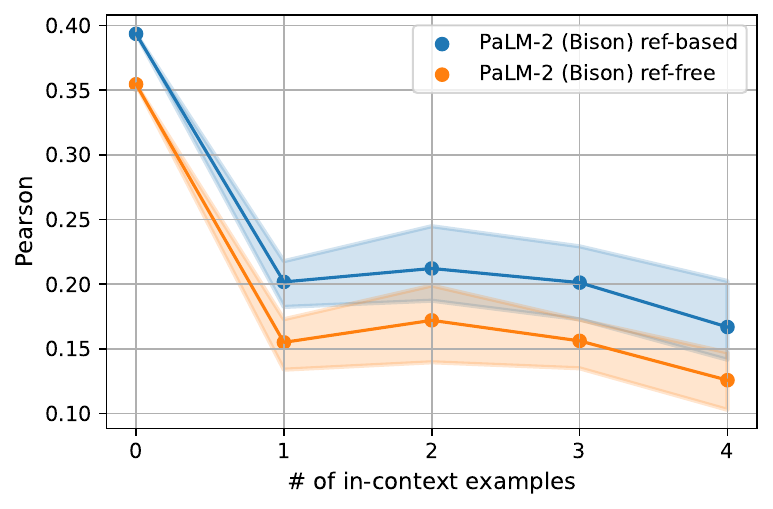}
    \vspace{-0.25em}
    \caption{\small Mean \textit{Pearson} and its interquartile range (IQR) in the WMT22 EN-DE test set, as we increase the number of in-context examples with \textit{stratified} sampling}
    \label{fig:kshot_performance}
    \vspace{-0.7em}
\end{figure}

\autoref{fig:kshot_performance} shows the mean and interquartile range (IQR) of the performance as we increase the number of in-context examples $k$ (with 100 example sets per $k$) sampled with \textit{stratified} sampling (see \autoref{app:additional-results} for \textit{uniform}). Surprisingly, despite evidence of the benefits of in-context learning for many tasks, we found that including in-context examples during evaluation (almost) never led to better performance, either with \textit{uniform} or \textit{stratified} sampling.

\begin{figure}[ht]
\centering
 \hspace{-1em}
\begin{subfigure}[b]{\linewidth}
\centering
\includegraphics[width=0.9\textwidth]{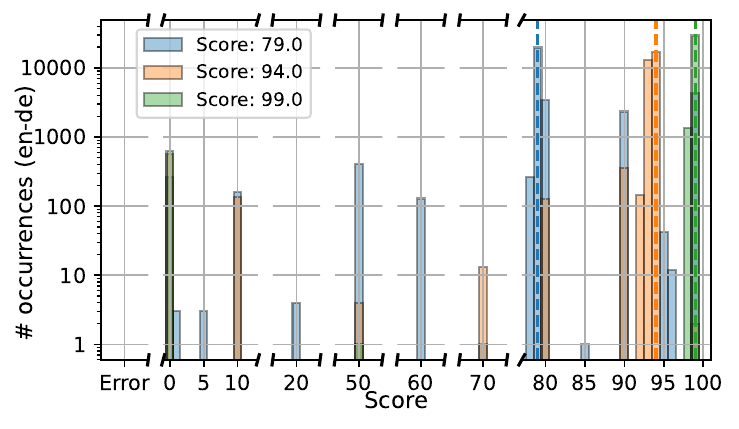}
\end{subfigure}
\\
  \hspace{-1em}
\begin{subfigure}[b]{\linewidth}
\centering
\includegraphics[width=0.9\textwidth]{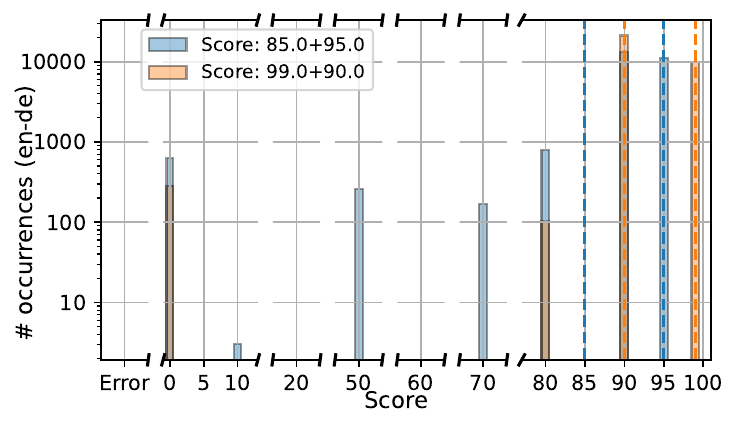}
\end{subfigure}
\vspace{-0.6em}
\caption{\small Distribution of scores for PaLM-2 (\bison{}) models for 1-shot (top) and 2-shot (bottom) setups, with various in-context learning sets for each (and their scores in the legend)}
\label{fig:distribution_of_scores_kshot}
\vspace{-0.2em}
\end{figure}

To investigate the cause of this disappointing performance, we looked at how \textit{particular} in-context example sets affect the distribution of scores produced by LLM-based evaluators. \autoref{fig:distribution_of_scores_kshot} shows the distribution of scores \textit{over the whole test set} for the 1-shot and 2-shot settings, with different in-context examples sets. We can see that output distribution is heavily biased by the scores in the in-context examples: despite \textit{never} predicting 79 in the zero-shot setting, when a single example with that score is included, it starts to dominate the model predictions. This seems to hint that LLMs ``overfit'' to the specific scores provided as examples, rather than generalizing to the broader evaluation task, which could explain the lackluster performance of in-context learning.

\subsection{Low Resource Languages} 
\autoref{tab:low-resource} shows the performance of PaLM-2 models at \textit{score prediction} for \textit{low-resource} translation. Overall, we find that similar to high-resource LPs, these models are good zero-shot evaluators, with system-level accuracies around 90\%. However, \textit{zero-shot} LLMs underperform \textit{learned} metrics, even when these metrics also weren't exposed to data in these low-resource languages.

\begin{table}[ht]
\centering
\resizebox{\linewidth}{!}{%
\setlength{\tabcolsep}{1pt}
\renewcommand{\arraystretch}{0.9}
\begin{tabular}{lcccccccccccccc}
\toprule
 &  & \multicolumn{1}{c}{\textbf{System}} & \multicolumn{4}{c}{\textbf{Segment} $\rho$} \\
\cmidrule{3-3}\cmidrule{4-7}
Model & Ref? & All & {\small EN-KK} & {\small EN-GU} & {\small KK-EN} & {\small GU-EN} \\
\midrule
\multicolumn{1}{@{}l}{\raisebox{1.2ex}{\scriptsize \textbf{Baseline}}}\\[-6pt]
MetricX-XXL$^\star$ & \cmark & 94.0\% & 0.666 & 0.701 & 0.539 & 0.409 \\
\midrule
\multicolumn{1}{@{}l}{\raisebox{1.2ex}{\scriptsize \textbf{Prompting}}}\\[-6pt]
\bison{} & \cmark & 92.2\% & 0.605 & 0.540 & 0.462 & 0.339 \\
\unicorn{} & \cmark & 87.4\% & 0.609 & 0.621 & 0.495 & 0.384 \\
\rowcolor{gray!08}\bison{} & \xmark & 89.8\% & 0.567 & 0.478 & 0.381 & 0.313 \\
\rowcolor{gray!08} \unicorn{} & \xmark & 84.4\% & 0.536 & 0.523 & 0.433 & 0.334 \\
\bottomrule
\end{tabular}%
}
\caption{\small Meta-evaluation results for system-level \textit{accuracy} and segment-level \textit{Pearson} on the low-resource languages, using PaLM-2 for \textit{score prediction}. $^\star$Note that the baseline is slightly different from the high-resource case, being trained on the same data but \textit{without} these \textit{low-resource} language pairs.}
\vspace{-1em}
\label{tab:low-resource}
\end{table} 

\begin{figure*}[ht]
    \centering
    \includegraphics[width=0.9\textwidth]{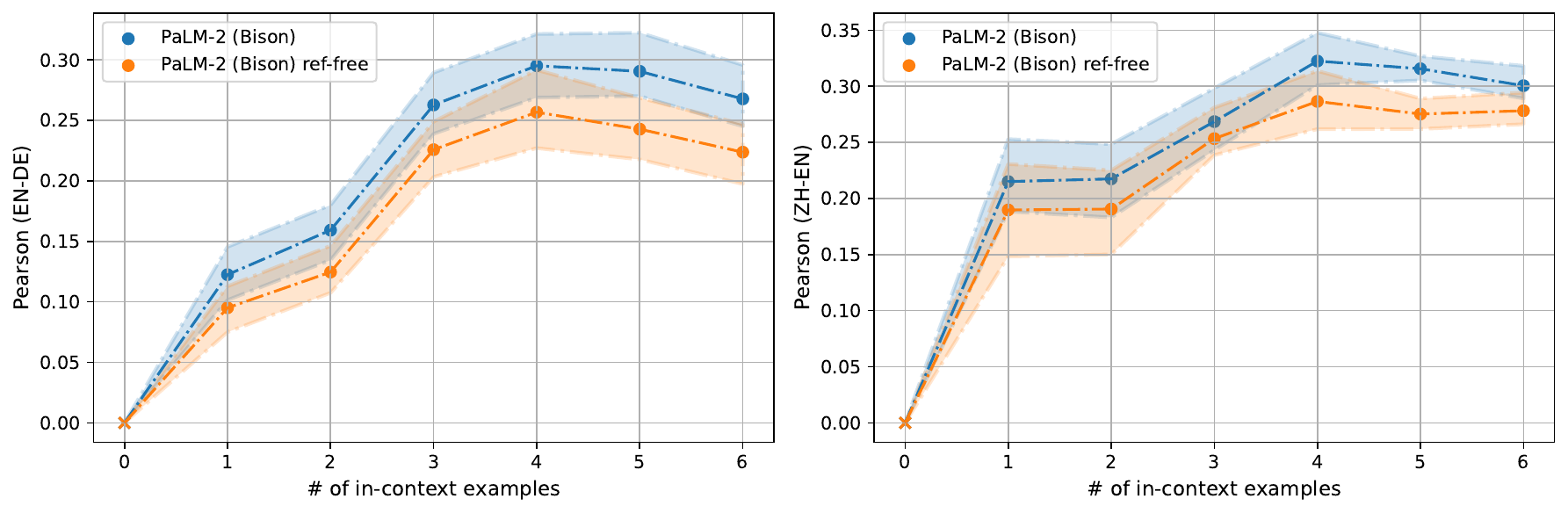}
    \vspace{-0.8em}
    \caption{Mean \textit{Pearson} and its interquartile range (IQR), as we increase the number of in-context examples in the \textsc{AutoMQM} prompt, for EN-DE (left) and ZH-EN (right).}
    \vspace{-0.3em}
    \label{fig:kshot_performance_automqm}
\end{figure*}

\subsubsection{\textsc{AutoMQM}}

\begin{table*}[ht]
\small
\centering
{\renewcommand{\arraystretch}{0.8}
\begin{tabular}{@{\extracolsep{2pt}}lcccccccccc}
\toprule
 &  & \multicolumn{1}{c}{\textbf{System-Level}} & \multicolumn{4}{c}{\textbf{Segment-Level}} \\
\cmidrule{3-3}\cmidrule{4-7}
 &  & All (2 LPs) & \multicolumn{2}{c}{EN-DE} & \multicolumn{2}{c}{ZH-EN} \\
\cmidrule{3-3}\cmidrule{4-5} \cmidrule{6-7}
Model & Ref? & Accuracy & $\rho$ & $\mathrm{acc}^\star$ & $\rho$ & $\mathrm{acc}^\star$ \\
\midrule
\multicolumn{1}{@{}l}{\raisebox{1.2ex}{\scriptsize \textbf{Baselines}}}\\[-6pt]
MetricX-XXL & \cmark & 81.1\% & 0.549 & 61.1\% & 0.581 & 54.6\% \\
\textsc{MaTESe} & \cmark & 79.9\% & 0.391 & 58.8\% & 0.528 & 51.5\% \\
\rowcolor{gray!8} COMET-QE & \xmark & 76.9\% & 0.419 & 56.3\% & 0.505 & 48.8\% \\
\rowcolor{gray!8} \textsc{MaTESe}-QE & \xmark & 73.4\% & 0.298 & 57.9\% & 0.468 & 50.1\% \\
\rowcolor{gray!8} \textsc{COMET-WL} & \xmark & 71.6\% & 0.418 & 57.1\% & 0.406 & 51.5\% \\
\midrule
\multicolumn{1}{@{}l}{\raisebox{1.2ex}{\scriptsize \textbf{Score Prediction}}}\\[-6pt]
PaLM-2 \bison{} & \cmark & 86.4\% & 0.394 & 56.8\% & 0.322 & 49.3\% \\
PaLM-2 \unicorn{} & \cmark & 86.4\% & 0.401 & 56.3\% & 0.349 & 51.1\% \\
\rowcolor{gray!8} PaLM-2 \bison{} & \xmark & 84.0\% & 0.355 & 57.0\% & 0.299 & 48.6\% \\
\rowcolor{gray!8} PaLM-2 \unicorn{} & \xmark & 80.5\% & 0.275 & 56.1\% & 0.252 & 48.3\% \\
\midrule
\multicolumn{1}{@{}l}{\raisebox{1.2ex}{\scriptsize \textbf{AutoMQM}}}\\[-6pt]
PaLM-2 \bison{} & \cmark & 84.0\% & 0.369 & 59.2\% & 0.355 & 48.4\% \\
PaLM-2 \unicorn{} & \cmark & 87.6\% & 0.432 & 59.1\% & 0.442 & 51.8\% \\
\rowcolor{gray!8} PaLM 2 \bison{} & \xmark & 87.6\% & 0.297 & 55.2\% & 0.331 & 48.0\% \\
\rowcolor{gray!8} PaLM 2 \unicorn{} & \xmark & 83.4\% & 0.368 & 56.4\% & 0.429 & 50.2\% \\
\bottomrule
\end{tabular}
\caption{Meta-evaluation results for PaLM-2 models using \textit{AutoMQM} and score prediction, at the system and segment levels for multiple language pairs.}
\label{tab:automqm}
}
\vspace{-1em}
\end{table*}

\begin{figure}[ht]
    \centering
    \includegraphics[width=0.95\linewidth]{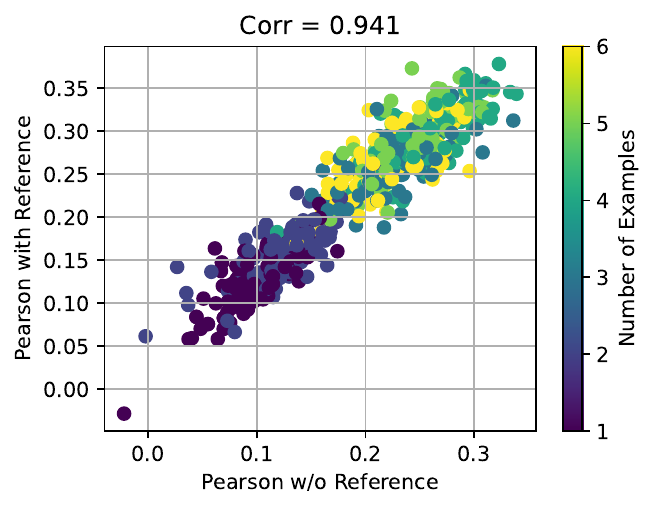}
    \vspace{-0.8em}
    \caption{\small Scatter plot of the \textit{Pearson} of PaLM-2 (\bison{}) models, with/without including the \textit{reference} in the prompt, for each in-context learning setting tried. }
    \label{fig:ref_based_free_correlations}
    \vspace{-1.2em}
\end{figure}

\autoref{fig:kshot_performance_automqm} shows the mean and interquartile range (IQR) of the performance of PaLM-2 \bison{} with \textsc{AutoMQM}, as we increase the number of in-context examples (again, with 100 example sets per $k$). Contrary to the performance with score prediction, we find that performance with \textsc{AutoMQM} seems to (mostly) scale with the number of in-context examples: performance increases monotonically with up to 4 in-context examples and plateaus thereafter. Additionally, the variance across the in-context learning sets seems to be lower, with most example sets exhibiting less than 0.05 \textit{Pearson} difference from the best-performing sets. All this suggests that LLM evaluators are much more robust to the choice of in-context examples when prompted for \textsc{AutoMQM} rather than for score prediction. We also find that the behavior of in-context learning is quite similar for both reference-based and reference-less evaluation tasks. Finally, we observe that the example sets that perform well for one task generally work well for the other, with performance on both settings given a fixed in-context set being highly correlated, as shown in \autoref{fig:ref_based_free_correlations}.

\begin{figure}[ht]
  \centering
  \hspace{-1em}
    \includegraphics[width=0.97\linewidth]{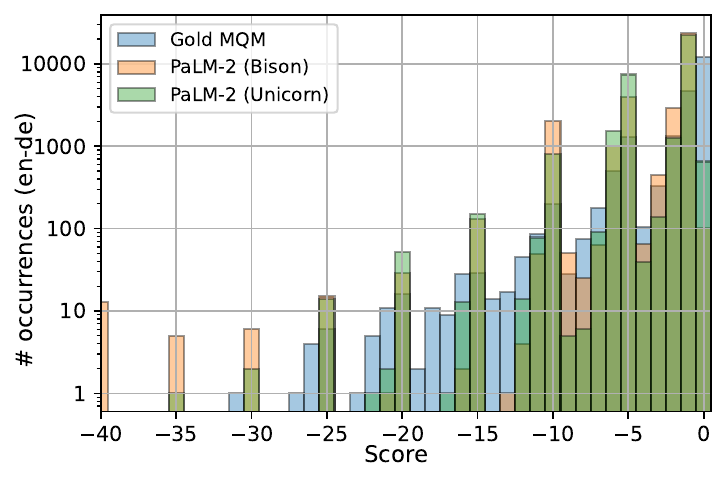}
  \vspace{-0.5em}
  \caption{\small Distribution of scores for PaLM-2 models using \textsc{AutoMQM}, on WMT22 EN-DE}
  \label{fig:distribution_of_scores_automqm}
  \vspace{-0.2em}
\end{figure}

\autoref{tab:automqm} shows the meta-evaluation results for PaLM-2 \bison{} and \unicorn{} prompted with \textsc{AutoMQM} (using the best-performing in-context learning sets in \autoref{fig:kshot_performance_automqm}). For ease of comparison, we also report their performance when prompted for \textit{score prediction}, as well as the performance of the baselines. Overall, prompting LLMs with \textsc{AutoMQM} seems to lead to significant improvements in evaluating machine translation quality, particularly for larger models: \unicorn{} achieves better performance (across all meta evaluations) with it than when prompted for \textit{score prediction}, and its reference-less version is competitive with the best learned metric even at the segment level. However, for the smaller \bison{}, the benefits of \textsc{AutoMQM} are less clear, with both techniques performing comparably. This hints that \textit{scale} is necessary for \textit{zero-} and \textit{few-} shot fine-grained evaluation (like with \textsc{AutoMQM}). We also find that the \textit{distribution} of scores produced by LLMs prompted with \textsc{AutoMQM} is much closer to the gold MQM distribution, with models outputting a much larger set of scores, and in the same ranges as annotators do (see \autoref{fig:distribution_of_scores_automqm}).

\begin{table}[ht]
\centering
\resizebox{\linewidth}{!}{%
\setlength{\tabcolsep}{1pt}
\renewcommand{\arraystretch}{0.9}
\begin{tabular}{@{\extracolsep{0.2pt}}lccccccc}
\toprule
 &  & \multicolumn{3}{c}{EN-DE} & \multicolumn{3}{c}{ZH-EN} \\
\cmidrule{3-3}\cmidrule{3-5} \cmidrule{6-8}
Model & R? & SP & MR & MCC & SP & MR & MCC \\
\midrule
\multicolumn{1}{@{}l}{\raisebox{1.2ex}{\small \textbf{Baselines}}}\\[-7pt]
\rowcolor{gray!8} COMET-WL & \xmark & 0.267 & 0.250 & 0.161 & 0.364 & 0.178 & 0.152 \\
\midrule
\multicolumn{1}{@{}l}{\raisebox{1.2ex}{\small \textbf{AutoMQM}}}\\[-7pt]
\bison{} & \cmark & 0.095 & 0.749 & 0.060 & 0.252 & 0.255 & 0.109 \\
\unicorn{} & \cmark & 0.175 & 0.628 & 0.193 & 0.238 & 0.476 & 0.143 \\
\rowcolor{gray!8} \bison{} & \xmark & 0.119 & 0.520 & 0.092 & 0.224 & 0.311 & 0.091 \\
\rowcolor{gray!8} \unicorn{} & \xmark & 0.150 & 0.580 & 0.150 & 0.229 & 0.488 & 0.133 \\
\bottomrule
\end{tabular}%
}
\caption{\small Span-level meta-evaluation on WMT22 for PaLM-2 models using \textit{AutoMQM}. \textbf{SR} and \textbf{MR} represent \textit{span precision} and \textit{major recall}, respectively.}
\label{tab:automqm_spanbased}
\end{table} 

Finally, when evaluating the error spans produced by LLMs prompted with \textsc{AutoMQM} (\autoref{tab:automqm_spanbased}), we find that PaLM-2 models are able to identify most of the \textit{major} errors. However, it does seem to \textit{over-predict} errors (with errors predicted by \unicorn{} having on average $\sim$5 words per span vs $\sim$2 words in the ground truth) and have overall low span precision. Similarly to overall \textit{score} correlations, \textit{scale} also seems to be important for the quality of spans produced by \textsc{AutoMQM}, with \unicorn{} outperforming \bison{} at most metrics. Additionally, \unicorn{} prompted with AutoMQM predicts spans of comparable quality to the ones produced by current state-of-the-art \textit{learned} word-level evaluators (trained on a considerable number of fine-grained annotations derived from MQM): while word-level models are more precise, their overall span correlation (MCC) is comparable, and they miss considerably more \textit{major} errors than LLMs (despite only leveraging a handful of annotations).



\section{Conclusion}

In this study, we have systematically investigated the capabilities of large language models for machine translation evaluation through \textit{score prediction}, and proposed \textsc{AutoMQM}, a novel prompting technique that leverages the Multidimensional Quality Metrics (MQM) framework for interpretable MT evaluation using LLMs.

We demonstrated that just prompting LLMs for score prediction leads to state-of-the-art system-level evaluators, but still falls short of the best \textit{learned} metrics at the segment-level (with finetuning being necessary to close this gap). Then we showed that \textsc{AutoMQM} can further improve the performance of LLMs without finetuning while providing interpretability through error spans that align with human annotations. 

Our findings surrounding finetuning LLMs for \textit{score prediction} hint that LLMs' performance in machine translation evaluation could be further improved by finetuning these models on fine-grained human judgment data (like MQM) and is a direction we are actively pursuing. Additionally, the general-purpose nature of LLMs may enable the application of similar prompting techniques (leveraging some fine-grained evaluation schemes) to other evaluation problems \cite{wu2023finegrained}.

\section*{Acknowledgements}

We would like to thank Ricardo Rei, Marcos Treviso and Chryssa Zerva for helping run the word-level QE baselines, and George Foster who provided feedback on an earlier version of this work.
This work was partially supported by EU's Horizon Europe Research and Innovation Actions (UTTER, contract 101070631), the P2020 program MAIA (LISBOA-01-0247-FEDER-045909), the Portuguese Recovery and Resilience Plan, and the Funda\c{c}\~ao para a Ci\^encia e Tecnologia through contracts SFRH/BD/150706/2020 and UIDB/50008/2020.

\bibliography{tacl, anthology}
\bibliographystyle{acl_natbib}

\iftaclpubformat

\onecolumn

\appendix

\section{Multidimensional Quality Metric (MQM)}
\label{app:mqm_description}

The Multidimensional Quality Metrics (MQM) framework is a flexible human-evaluation framework developed to evaluate and categorize errors in translations. Annotators are instructed to identify all 
errors within each segment in a document, paying particular attention to document context. See \autoref{tab:mqm-guidelines} for the annotator guidelines provided.

\begin{table*}[!htb]\centering
\scalebox{1.00}{
\noindent\fbox{%
\parbox{1.0\textwidth}{%

You will be assessing translations at the segment level, where a segment may contain one or more sentences. Each segment is aligned with a corresponding source segment, and both segments are displayed within their respective documents. Annotate segments in natural order, as if you were reading the document. You may return to revise previous segments.\\

Please identify all errors within each translated segment, up to a maximum of five. If there are more than five errors, identify only the five most severe. If it is not possible to reliably identify distinct errors because the translation is too badly garbled or is unrelated to the source, then mark a single \emph{Non-translation} error that spans the entire segment.\\
 
To identify an error, highlight the relevant span of text, and select a category/sub-category and severity level from the available options. (The span of text may be in the source segment if the error is a source error or an omission.) When identifying errors, please be as fine-grained as possible. For example, if a sentence contains two words that are each mistranslated, two separate mistranslation errors should be recorded. If a single stretch of text contains multiple errors, you only need to indicate the one that is most severe. If all have the same severity, choose the first matching category listed in the error typology (eg, \emph{Accuracy}, then \emph{Fluency}, then \emph{Terminology}, etc).\\
 
Please pay particular attention to document context when annotating. If a translation might be questionable on its own but is fine in the context of the document, it should not be considered erroneous; conversely, if a translation might be acceptable in some context, but not within the current document, it should be marked as wrong.\\
 
There are two special error categories: \emph{Source error} and \emph{Non-translation}. Source errors should be annotated separately, highlighting the relevant span in the source segment. They do not count against the five-error limit for target errors, which should be handled in the usual way, whether or not they resulted from a source error. There can be at most one \emph{Non-translation} error per segment, and it should span the entire segment. No other errors should be identified if \emph{Non-Translation} is selected.}
}}
\vspace{-1em}
\caption{MQM annotator guidelines}
\vspace{-1em}
\label{tab:mqm-guidelines}
\end{table*}

Annotators are asked to assign both an error \textit{severity} and \textit{category}. Error \textit{severity} (either \textit{major} or \textit{minor}) is assigned independently of category. Spans with no marked errors have \textit{neutral} severity and no category. Possible error categories are displayed in \autoref{tab:mqm-hierarchy}. 

\begin{table*}\centering
\scalebox{0.80}{
\begin{tabular}{ll|l}\toprule
\multicolumn{2}{l|}{Error Category} & Description \\
\midrule
Accuracy & Addition    & Translation includes information not present in the source. \\
    & Omission         & Translation is missing content from the source. \\
    & Mistranslation   & Translation does not accurately represent the source.\\
    & Untranslated text & Source text has been left untranslated. \\
\midrule
Fluency & Punctuation   & Incorrect punctuation (for locale or style). \\
    & Spelling          & Incorrect spelling or capitalization. \\
    & Grammar           & Problems with grammar, other than orthography. \\
    & Register          & Wrong grammatical register (eg, inappropriately informal pronouns). \\
    & Inconsistency     & Internal inconsistency (not related to terminology). \\
    & Character encoding          & Characters are garbled due to incorrect encoding. \\
\midrule
Terminology & Inappropriate for context & Terminology is non-standard or does not fit context.\\
            & Inconsistent use & Terminology is used inconsistently.\\
\midrule
Style & Awkward & Translation has stylistic problems.\\
\midrule
Locale & Address format & Wrong format for addresses.\\
convention  & Currency format & Wrong format for currency.\\
    & Date format & Wrong format for dates. \\
    & Name format & Wrong format for names. \\
    & Telephone format & Wrong format for telephone numbers. \\
    & Time format & Wrong format for time expressions. \\
\midrule
Other & & Any other issues. \\
\midrule
Source error & & An error in the source. \\
\midrule
Non-translation & & Impossible to reliably characterize distinct errors.\\
\bottomrule
\multicolumn{3}{c}{}\\
\end{tabular}
}
\vspace{-1em}
\caption{MQM hierarchy.}
\label{tab:mqm-hierarchy}
\end{table*}

Since MQM doesn't ask annotators for quality scores, those scores are derived automatically from the identified error spans and their classifications, based on a \textit{weighting} of each error severity and category. \autoref{tab:mqm-scoring} summarizes this weighting scheme, in which segment-level scores can range from 0 (perfect) to 25 (worst). The final segment-level score is an average over scores from all annotators. In some settings (e.g. calculating correlation for learned metrics), the scores are negated.

\begin{table}\centering
\begin{tabular}{l|l|l}\toprule
Severity & Category & Weight \\
\midrule
Major & Non-translation & 25 \\
      & all others & 5 \\
\midrule
Minor & Fluency/Punctuation & 0.1 \\
      & all others          & 1 \\
\midrule
Neutral & all & 0 \\
\bottomrule
\end{tabular}
\caption{MQM error weighting.}
\label{tab:mqm-scoring}
\end{table}

We use the same weighting to obtain scores from errors identified by \textsc{AutoMQM}.

\section{Sampling in-context learning examples for AutoMQM}
\label{app:rejection-criteria}

\autoref{fig:rejection-criteria} shows the rejection criteria used when sampling example sets as discussed in \S\ref{sec:score_prediction}.

\begin{figure}
    \centering
\begin{lstlisting}[language=Python, style=codestyle]
def check_icl_set(
    examples: pd.DataFrame,
    min_errors=3,
    majmin_threshold=2,
    cat_diversity=2,
    min_clen=20,
    max_clen=400,
):
    # Check if they have the same number of spans as severity/category
    if not examples.apply(
        lambda r:
          len(r['span']) == len(r['severity']) and len(r['span']) == len(r['category']),
        axis=1
    ).all():
      return False

    # Check if there are at least min_errors
    if examples['severity'].apply(lambda svs: len(svs)).sum() < min_errors:
      return False

    # Check that there's a balance of major and minor errors.
    major_count = examples['severity'].apply(lambda svs: sum([s=='major' for s in svs])).sum()
    minor_count = examples['severity'].apply(lambda svs: sum([s=='minor' for s in svs])).sum()
    if abs(major_count - minor_count) > majmin_threshold:
        return False

    # Check that at least cat_diversity error types are represented.
    categories = examples['category'].apply(lambda cs: [c.split("/")[0] for c in cs])
    represented_error_types = set().union(*categories.tolist())
    if len(represented_error_types) < cat_diversity:
        return False

    top_clen = examples.apply(
        lambda row: max(len(row[s]) for s in ('source', 'reference', 'candidate')
    ), axis=1).max()
    bot_clen = examples.apply(
        lambda row: min(len(row[s]) for s in ('source', 'reference', 'candidate')), 
    axis=1).min()

    if top_clen > max_clen or bot_clen < min_clen:
      return False

    # All checks passed.
    return True
\end{lstlisting}
    \caption{Rejection criteria used when sampling \textit{in-context learning} examples for \textsc{AutoMQM}.}
    \label{fig:rejection-criteria}
\end{figure}

\section{Additional Results}
Figures \ref{fig:kshot_performance_ende}, \ref{fig:kshot_performance_zhen}, and \ref{fig:distribution_of_scores_zhen} present additional experimental results.
\label{app:additional-results}

\begin{figure*}[ht]
    \centering
    \includegraphics[width=0.9\textwidth]{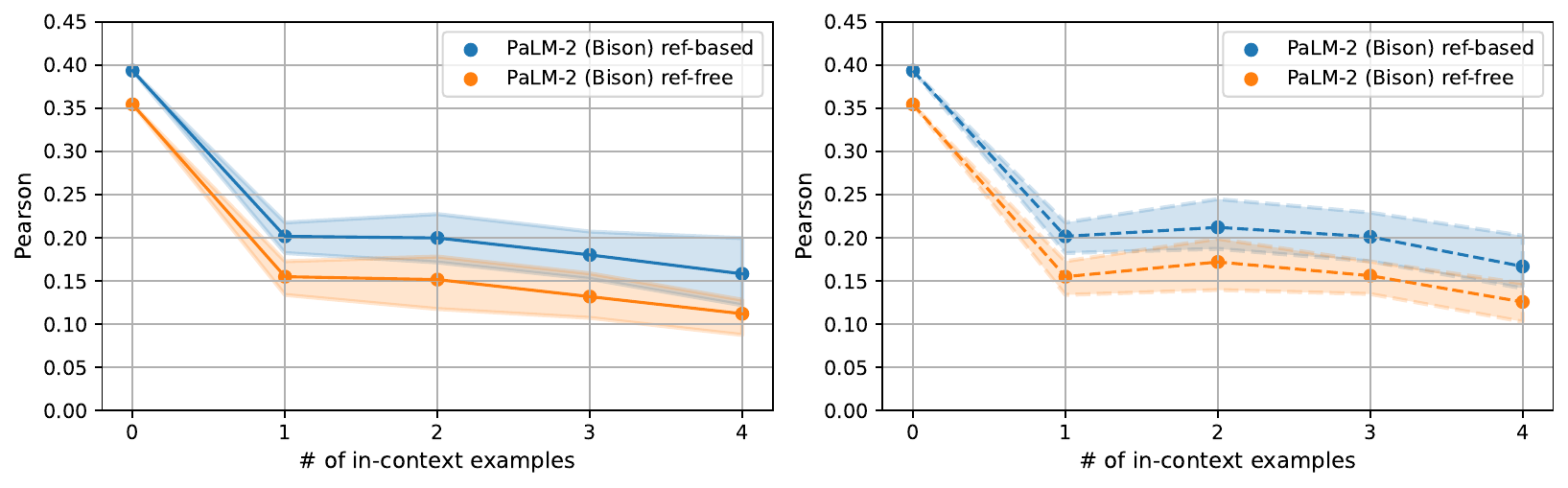}
    \caption{Mean \textit{Pearson} and its interquartile range (IQR), as we increase the number of in-context examples in the \textit{score prediction} prompt, sampled with \textit{uniform} (left) and \textit{stratified} (right) sampling, for WMT22 EN-DE.}
    \label{fig:kshot_performance_ende}
\end{figure*}

\begin{figure*}[ht]
    \centering
    \includegraphics[width=0.9\textwidth]{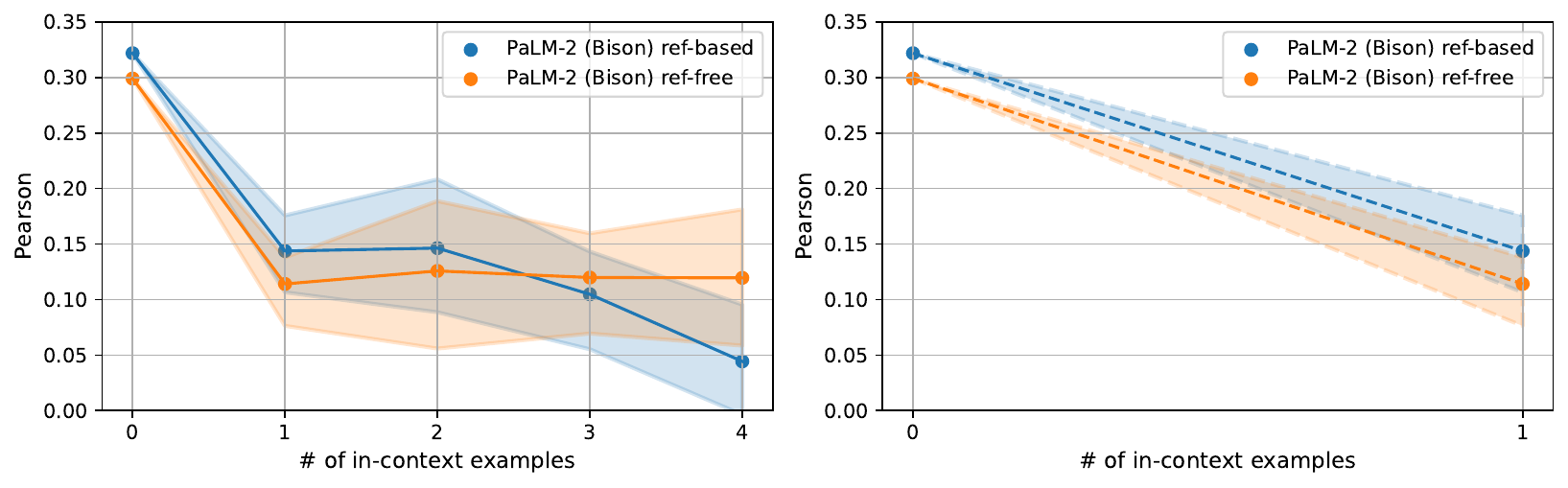}
    \caption{Mean \textit{Pearson} and its interquartile range (IQR), as we increase the number of in-context examples in the \textit{score prediction} prompt, sampled with \textit{uniform} (left) and \textit{stratified} (right) sampling, for WMT22 ZH-EN.}
    \label{fig:kshot_performance_zhen}
\end{figure*}

\begin{figure}[ht]
  \centering
    \includegraphics[width=0.5\linewidth]{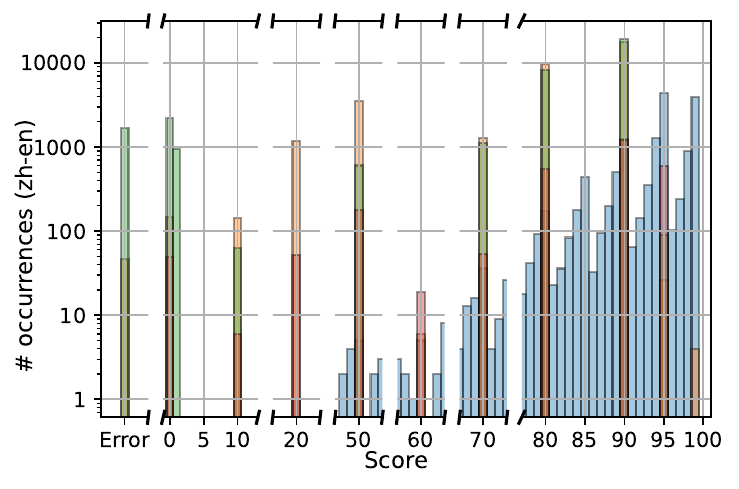}
  \vspace{-1em}
  \caption{\small Distribution of scores for various LLM \textit{reference-based} evaluators, on the ZH-EN test set. Note that the $y$ axis is in \textit{log-scale}.}
  \label{fig:distribution_of_scores_zhen}
  \vspace{-1.2em}
\end{figure}

\fi

\end{document}